\documentclass[journal]{IEEEtran}  

\usepackage{graphics} 
\usepackage{epsfig} 
\usepackage{float}
\usepackage[hidelinks]{hyperref}
\usepackage{siunitx}
\usepackage{amsmath} 
\usepackage{amssymb}  
\usepackage{graphicx} 
\usepackage{cite}
\usepackage{algorithm,algorithmic}
\usepackage{booktabs}
\usepackage{multirow}
\usepackage{dblfloatfix}
\usepackage[utf8]{inputenc}
\usepackage[T1]{fontenc}

\usepackage{nomencl}
\makenomenclature

\begin{document}

\title{A Novel Tropical Geometry-based Interpretable Machine Learning Method: Application in Prognosis of
Advanced Heart Failure
}

\author{Heming Yao$^{1}$, Harm Derksen$^{2}$, Jessica R. Golbus$^{3}$, Justin Zhang$^{4}$, Keith D. Aaronson$^{3}$, Jonathan Gryak$^{1,5}$, and Kayvan Najarian$^{1,5,6,7}$
\thanks{$^{1}$Department of Computational Medicine and Bioinformatics, University of Michigan, Ann Arbor, MI, USA}%
\thanks{$^{2}$Department of Mathematics, Northeastern University, Boston, MA, USA}
\thanks{$^{3}$Division of Cardiovascular Medicine, Department of Internal Medicine, University of Michigan, Ann Arbor, MI, USA}%
\thanks{$^{4}$Electrical and Computer Engineering, College of Engineering, University of Michigan, Ann Arbor, MI, USA}%
\thanks{$^{5}$Michigan Institute for Data Science, University of Michigan, Ann Arbor, MI, USA}%
\thanks{$^{6}$Department of Emergency Medicine, University of Michigan, Ann Arbor, MI, USA}%
\thanks{$^{7}$Michigan Center for Integrative Research in Critical Care, University of Michigan, Ann Arbor, MI, USA}%
}

\maketitle


\begin{abstract}
A model's interpretability is essential to many practical applications such as clinical decision support systems. In this paper, a novel interpretable machine learning method is presented, which can model the relationship between input variables and responses in humanly understandable rules. The method is built by applying tropical geometry to fuzzy inference systems, wherein variable encoding functions and salient rules can be discovered by supervised learning. Experiments using synthetic datasets were conducted to investigate the performance and capacity of the proposed algorithm in classification and rule discovery. Furthermore, the proposed method was applied to a clinical application that identified heart failure patients that would benefit from advanced therapies such as heart transplant or durable mechanical circulatory support. Experimental results show that the proposed network achieved great performance on the classification tasks. In addition to learning humanly understandable rules from the dataset, existing fuzzy domain knowledge can be easily transferred into the network and used to facilitate model training. From our results, the proposed model and the ability of learning existing domain knowledge can significantly improve the model generalizability. The characteristics of the proposed network make it promising in applications requiring model reliability and justification. 

\end{abstract}

\IEEEpeerreviewmaketitle

\section{INTRODUCTION}

Heart failure (HF) afflicts 6.5 million Americans 20 and older, with its prevalence projected to increase annually \cite{parikh2018heart, benjamin137american}. Treatment of these patients remains limited both by medical therapies and by organ availability. The appropriate delivery of advanced therapies, heart transplantation (HT) or mechanical circulatory support (MCS) implantation, to patients with end-stage HF is highly nuanced and requires expertise from advanced HF cardiologists. Due to the high prevalence of HF, the majority of patients are managed by primary care physicians or cardiologists, who lack training in the management of these patients. Thus, there is a need for artificial intelligence (AI) based tools that can systematically identify patients warranting referral to an advanced HF cardiologist for consideration of HT or MCS implantation. In this study, we aims to build a clinical decision-making model that can differentiate patients eligible for and most likely to benefit from advanced therapies; such as durable MCS or HT; from those too well, too sick, or otherwise ineligible for advanced therapies. 

AI and machine learning (ML) have been increasingly applied to healthcare problems \cite{noorbakhsh2019artificial}. Previous studies investigated AI in disease diagnosis, treatment effectiveness prediction, and patient outcome prediction \cite{caccomo2018fda, myszczynska2020applications, senders2018machine, yao2020automated}. Several studies have shown that AI performs as well as or better than humans \cite{davenport2019potential}. With a lower cost, AI-based decision support systems have the potential to improve patient management.

Despite tremendous progress in the field of AI/ML-based clinical decision support systems, there are still significant challenges that prevent widespread use of these methods in sensitive applications. While traditional models such as linear models and decision trees provide accessible reasoning, they are less capable of achieving high performance on complicated clinical problems. In contrast, a wide spectrum of ML models with higher complexity, including families of neural networks and support vector machines (SVM), can yield good metrics on experimental datasets. However, these “black box” models lack transparency and justification of their recommendations, making them much less likely to be trusted in clinical applications. Moreover, many popular ML methods, such as deep learning, utilize a large number of parameters, thus requiring large training datasets to avoid overfitting the data. However, in many clinical applications, collecting large annotated training datasets may be costly or even impossible. As such, there is a clear need for an interpretable ML model that can reliably model data using relatively small training sets. In addition, in healthcare applications, there exist many invaluable heuristics derived from domain knowledge expertise, often in the form of approximate rules that are used by human experts. For example, when caring for patients with end-stage HF, cardiologists use their clinical intuitions, paired with transplant guidelines, to identify patients who may benefit from a durable MCS device or HT. In the majority of existing AI/ML models, there is no clear mechanism to leverage such approximate knowledge for model formation or training.

The motivation of this study is to solve the aforementioned limitations in the field of AI. An interpretable ML algorithm is proposed to produce a transparent classification model and leverage existing domain knowledge to improve model generalizability and reliability. The proposed network is built upon tropical geometry and fuzzy inference systems \cite{zadeh1975fuzzy,takagi1985fuzzy}, a type of approximate reasoning method that has been used for multidimensional system modeling \cite{chan2011diagnosis, zhang2017nonlinear}. In this study, an algorithm with adaptive fuzzy subspace division and rule discovery was developed. The input encoding functions and the aggregation operators in classical fuzzy inference networks were reformulated by introducing tropical geometry \cite{mikhalkin2006tropical}, a piecewise-linear version of conventional algebraic geometry. Two synthetic datasets and one practical application in clinical decision support for patients with advanced HF were investigated to demonstrate the capability and interpretability of the proposed model. 

Our contributions in this study can be summarized as: 

\begin{enumerate}
  \item A novel interpretable ML algorithm was proposed, whose resulting recommendations and predictions would be transparent to users such as clinicians and patients. The model can produce humanly understandable rules, enabling new clinical knowledge discovery. The proposed network was validated using synthetic data with ground truth reasoning and a dataset from patients with HF. The experimental results show that the network has the capability to extract hidden rules from datasets. In addition, the proposed network achieved comparable or better performance than other ML models.

  \item Using the proposed algorithm, approximate domain knowledge can be directly incorporated into model training. The existing domain knowledge can improve the model's performance and reduce the need for a large training set, which makes it particularly appropriate for clinical applications. From our experimental results, initializing a network with existing approximate knowledge can significantly improve the model's accuracy.
  
  \item The proposed ML algorithm was successfully used to identify patients with HF eligible for advanced therapies, a highly sensitive application in medicine. From our results, the proposed algorithm achieved a significantly smaller generalization error, especially when existing knowledge was integrated into the network. The rules from the trained network were visualized and validated by cardiologists. The developed model can improve care for patients with HF by providing assessments that can be used by general providers without HF expertise.
\end{enumerate}

\section{Related Work}
\label{sec:related_work}
\subsection{Interpretable ML models}
One of the most popular definitions of interpretability is “the ability to explain or to present in understandable terms to a human” \cite{doshi2017towards}. There are primarily two bodies of work related to model interpretability: post-hoc interpretation and transparency\cite{mittelstadt2019explaining}.

Post-hoc interpretation methods are dedicated to explaining pre-developed “black box” ML models. For example, the interpretability of a random forest model was investigated by measuring variable importance \cite{louppe2014understanding}. \cite{hohman2019s} proposed Local Interpretable Model-agnostic Explanations (LIME), which explains the individual predictions of any classifier by learning local surrogate models that approximate the predictions from the target “black box” model. In \cite{hohman2019s}, an attribution graph summarizes neuron associations that contribute to a model’s predictions. While post-hoc methods reveal how powerful models works, they are approximations and have limited capacity in elucidating how to further improve the model.

In contrast, transparency addresses how a model functions internally by its structure and can provide exact explanations. They are usually less accurate than powerful “black box” ML models. The simplest transparent models are linear models, but these may fail whenever the relationships between features and responses are non-linear. The Na\"ive Bayes classifier calculates the probability for a class depending upon the value of the feature so that the contribution of each feature is evident. Decision trees are another class of transparent models that can capture interactions among different features. However, the structure of the decision tree is quite unstable and highly dependent on feature selection for each split. Generalized additive models (GAMs) are extended linear models that can capture non-linear relationships between individual features (or pairwise interactions) and responses \cite{lou2012intelligible}. They have been used in practical applications and exhibit good performance and interpretability \cite{caruana2015intelligible}. However, they are less capable modeling in high-dimensional feature interactions. Another type of transparent model is a fuzzy inference model, which models the relationship between features and responses by constructing compositional rules \cite{zadeh1975fuzzy}. Fuzzy inference models are designed for problems with inherent imprecision and uncertainty. In fuzzy inference models, knowledge is represented in the format of fuzziness of antecedents, consequents, and relations. As rules closely approximate human logic in decision-making, and fuzziness often exists in practical applications and especially in healthcare, the proposed network in this study is designed to leverage fuzzy logic and inference systems.

\subsection{Fuzzy inference system}
Previous studies have shown that fuzzy inference systems can be used for non-linear system approximation and rule identification \cite{chan2011diagnosis, zhang2017nonlinear}. While decisions produced by conventional AI/ML models are often opaque, hindering knowledge extraction and transfer, fuzzy inference models can extract humanly understandable knowledge from data. Classical fuzzy inference models utilize membership functions such as triangular functions to transform crisp inputs to a membership degree of fuzzy concepts. After that, a set of concepts are aggregated by T-norm and T-conorm operators (aggregation operators) to construct if-then rules, with the crisp output from each rule then transformed into output. $\min$ (T-norm) and $\max$ (T-conorm) are commonly used operators in fuzzy logic \cite{zadeh1975fuzzy,stoica1996synaptic}. A wide spectrum of fuzzy inference systems utilize the Takagi-Sugeno (TS) inference model \cite{takagi1985fuzzy}, whereby a complete rough partition of the input space is generated and an input-output relation is formed for each subspace. Adaptive Network-based Fuzzy Inference System (ANFIS)\cite{jang1993anfis} is a hybrid of a feed-forward neural network and fuzzy inference system with supervised learning capability that can be used to update the input-output relation in each subspace. ANFIS has been successfully applied in multiple applications \cite{cabalar2012some, al2016application}. In our previous work \cite{yao2019using}, an adaptive fuzzy inference network was developed and optimized using a genetic algorithm to identify patients eligible for advanced therapies. From our results, the network achieved good classification performance and provided transparent rules. 

However, the designs of the TS model and ANFIS pose challenges in practical complex applications where the number of input variables is relatively large  as this results in exponential growth in the number of subspaces (as well as the number of parameters). To handle this problem, a flexible $k$-d tree \cite{sugeno1991successive} and quadtree \cite{sun1994rule} have been adopted for input space partition, but are limited in that it is more challenging to assign understandable terms to membership functions using these methods. In this study, unlike previous methods, we propose an end-to-end network that will adaptively and iteratively discover subspaces related to each class using gradient-based back-propagation.

\section{Method}
\label{sec:method}

\begin{figure*}[] 
\centering
\scalebox{0.80}{
\includegraphics[width=7in]{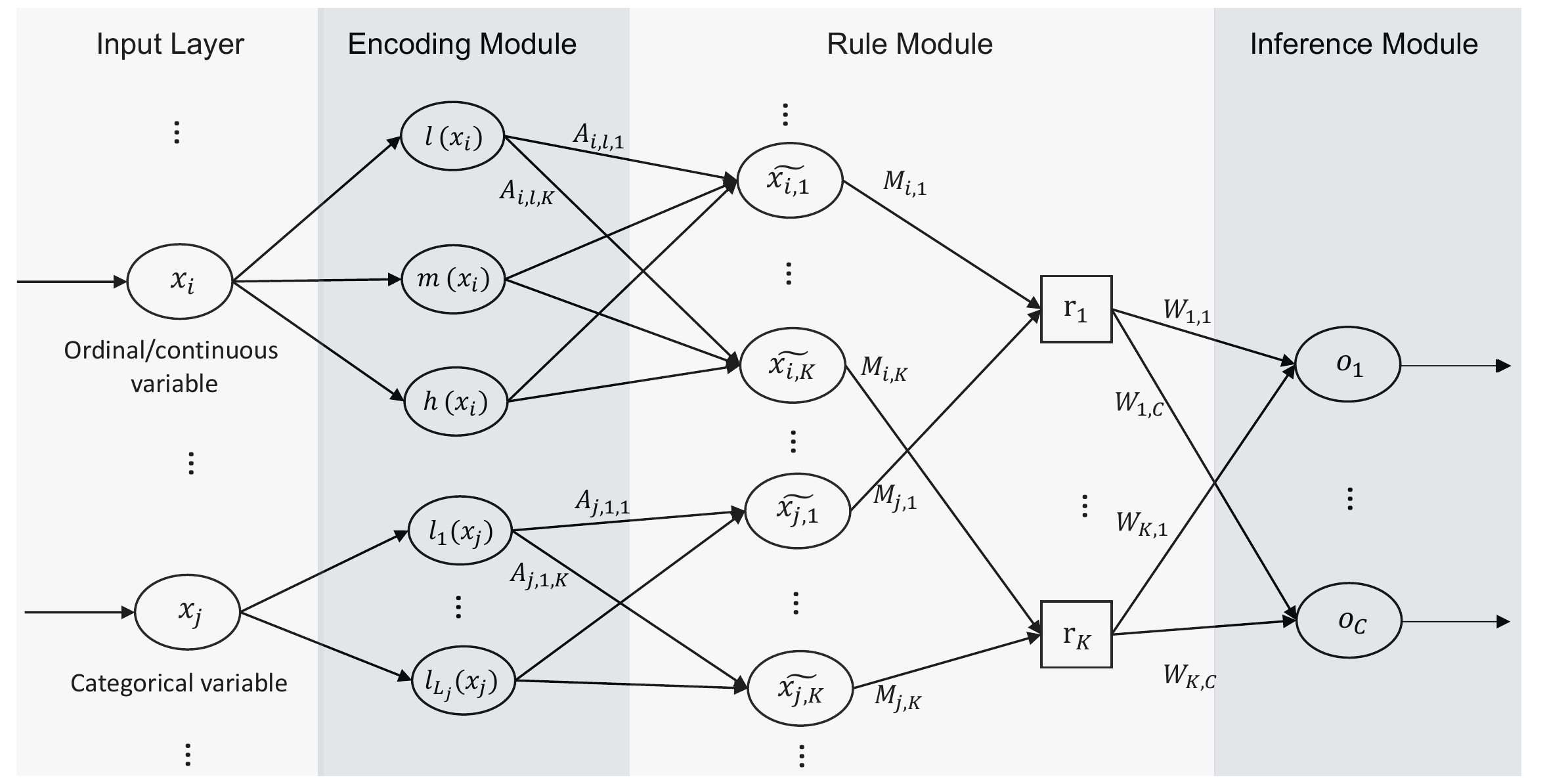}
}
\caption{An overview of the proposed network. The proposed network consists of an input layer, encoding module, rule module, and inference module. The nomenclatures we used in the diagram are described in Section \ref{sec:method}.}
\label{fig:overview}
\end{figure*}


\subsection{Overview of the proposed work}
In this study, a transparent end-to-end network was designed that can discover fuzzy subspaces contributing to each class. Figure \ref{fig:overview} depicts the proposed network. The proposed network has three major components:  an encoding module, a rule module, and an inference module. In the encoding module, input variables are encoded into humanly understandable fuzzy concepts. In the rule module, which contains trainable attention and connection matrices, a limited number of fuzzy subspaces (i.e., rules) are constructed as combinations of fuzzy concepts from the encoding module. Finally, by utilizing the inference matrix and the firing strength of each rule node, the probabilities of a sample belonging to each class are calculated in the inference module. In this network, parameters used for input encoding, subspace construction, and output inference are all trainable by gradient-based back-propagation. 

Unlike prior work on fuzzy inference systems, tropical geometry is used in this study to parametrize the aggregation operators and membership functions, with the parameter $\epsilon$ used to control their smoothness. Previously, $\min$ / $\mathrm{product}$ and $\max$ / $\mathrm{addition}$ were used as T-norm and T-conorm (i.e., as aggregation operations), respectively, though it remains unknown which of these operations is superior \cite{zadeh1975fuzzy,stoica1996synaptic,pratama2016incremental}. Similarly, it is unclear which membership function is optimal with respect to fuzzy set encoding, with triangular, trapezoidal, and Gaussian membership functions all commonly used. The use of Gaussian membership functions, \textrm{product}, and \textrm{addition} enable the application of back-propagation for optimization. However, it is unknown whether the lack of piecewise linearity limits the capability of a fuzzy inference system. In addition, while the selection of membership function shape may be application-specific, several prior studies have shown that the triangular membership function is superior to other membership functions \cite{ali2015comparison, monicka2011performance, sadollah2018fuzzy}. Previous studies also demonstrated that some practical problems are easier to solve in tropical geometry due to the piecewise linear nature of tropical objects \cite{mikhalkin2006tropical}. As such, parametrizing the membership functions, T-norm, and T-conorm allows the model to discover optimal encoding functions and operations during the training process. Throughout the course of the optimization process, these parametrized functions are gradually updated to be closer to piecewise linear functions, which both ensures the stability and convergence of gradient descent and results in an interpretable and accurate model. After model training, the attention matrix, connection matrix, and inference matrix can be used to interpret the model in the form of rules. 

As the proposed network mimics human logic, not only can knowledge be extracted from the trained model but also existing knowledge can be integrated/transferred into the model. In this study, experiments were performed to investigate whether initializing the network with existing domain knowledge can facilitate model training.

\subsection{Encoding module}
The input variables can be either ordinal, continuous, or categorical. For ordinal and continuous variables, fuzzy theory will be used to encode variables into multiple fuzzy sets. Unlike with crisp sets, for which membership is binary, for fuzzy sets a membership value in $[0,1]$ will be assigned to a variable's observed value for a given fuzzy set, indicating the confidence of that value belonging to the set. Fuzzy set membership approximates the fuzzy concept used by human experts during decision-making. For example, given the heart rate of a patient, the clinician may describe it as a “low” / “medium” / “high” heart rate. “Low”, “medium”, and “high” are the fuzzy concepts used in clinical problems. In this study, we encoded clinical ordinal/continuous variables into these three concepts. With an ordinal/continuous variable $x$, the membership functions $l(x), m(x), h(x)$ for “low”, “medium”, and “high” concepts are defined as

\begin{subequations}
\begin{align}
f_{\epsilon_1}(x) = &\epsilon_1 \log(1+\exp(x/\epsilon_1)),\\
l(x) = &f_{\epsilon_1}\left(\frac{a_{i,2}-x}{a_{i,2}-a_{i,1}}\right) - f_{\epsilon_1}\left(\frac{a_{i,1}-x}{a_{i,2}-a_{i,1}}\right),\\ 
m(x) = &f_{\epsilon_1}\left(\frac{x-a_{i,1}}{a_{i,2}-a_{i,1}}\right) - f_{\epsilon_1}\left(\frac{x-a_{i,2}}{a_{i,2}-a_{i,1}}\right) - \notag\\   
&f_{\epsilon_1}\left(\frac{a_{i,3}-x}{a_{i,4}-a_{i,3}}\right) + f_{\epsilon_1}\left(\frac{a_{i,4}-x}{a_{i,4}-a_{i,3}}\right) - 1, \\
h(x) = &f_{\epsilon_1}\left(\frac{x-a_{i,3}}{a_{i,4}-a_{i,3}}\right) - f_{\epsilon_1}\left(\frac{x-a_{i,4}}{a_{i,4}-a_{i,3}}\right), 
\end{align}
\label{eq:encoding}
\end{subequations}
where $a_{i,1}<a_{i,2}<a_{i,3}<a_{i,4}$ and are trainable. With $ 0 < \epsilon_1 < 1$, the membership functions are differentiable, with their smoothness modulated by $\epsilon_1$. As $\lim_{\epsilon_1 \to 0} f_{\epsilon_1}(x)=\max(0, x)$, when $\epsilon_1$ approaches 0, the membership functions in Equation \ref{eq:encoding} are close to trapezoidal membership functions or triangular membership functions (if $a_{i,2}$ is close to $a_{i,3}$). 

Using the defined membership functions, $x_i$ will be encoded as membership values in three fuzzy concepts: $l(x_i), m(x_i), h(x_i)$. In this study, we used three concepts - “low”, “medium”, and “high” - as they are commonly used in healthcare applications. The above formulations can be easily extended to a higher number of concepts.

Categorical variables are represented via a one-hot encoding directly and no fuzzy concepts are used. We denote $L_j$ as the number of levels of a categorical variable $x_j$. In this study, $x_j$ is encoded into $l_1(x_j), l_2(x_j), \ldots, l_{L_j}(x_j)$, where only one of them has a value of $1$ and all others are $0$.

\subsection{Rule module}
The rule module consists of two layers in the proposed architecture. In this module, the firing strength of a number of rules (fuzzy subspaces) are calculated for the classification task and denoted as $r_1, \ldots, r_K$ in Figure \ref{fig:overview}, where $K$ is the total number of rules.

\subsubsection{The first layer}
The first layer of the rule module selects the most relevant concept from each variable with respect to each rule using an attention matrix $\mathbf{A}$. $\mathbf{A}$ is the partitioned matrix formed by concatenating submatrices $\mathbf{A}_1, \mathbf{A}_2, \ldots, \mathbf{A}_H$, where $\mathbf{A}_h$ is the attention submatrix for the input variable $x_h$ and $H=I+J$ is the total number of input variables, with $I$ and $J$ the total number of ordinal/continuous and categorical variables, respectively. For an ordinal/continuous variable $x_i$, the submatrix $\mathbf{A}_i$ with entries $A_{i,m,n}$ has dimension $3\times K$, where $3$ is the number of concepts for ordinal/continuous variables used in this study and $K$ is the number of rules utilized in the network. For a categorical variable $x_j$, the submatrix $\mathbf{A}_j$ with entries $A_{j,m,n}$ has dimension $L_j\times K$. Thus, the attention matrix $A$ has dimension $(3I+\sum_j L_j)\times K$. 

For an ordinal/continuous variable $x_i$, the entry $A_{i,1,k}$ in the attention matrix represents the contribution of $x_i$ being “low” to rule $k$ (and similarly, $A_{i,2,k}$ for $x_i$ being “medium” and $A_{i,3,k}$ for $x_i$ being “high”). Entries in the attention matrix are all trainable and constrained to $[0, 1]$ by the hyperbolic tangent activation function. A higher value in $\mathbf{A}$ indicates a higher contribution. As shown in Figure \ref{fig:overview}, for an input variable $x_i$, the corresponding output from the first layer of the rule module is $\widetilde x_i$, a vector of length $K$. $\widetilde x_{i, k}$, the $k^{th}$ element of $\widetilde x_i$, is the firing strength of $x_i$ involved in $k^{th}$ rule. 

For an ordinal/continuous variable $x_i$ and categorical variable $x_j$, $\widetilde x_{i, k}$, and $\widetilde x_{j, k}$ are calculated as:

\begin{subequations}
\begin{align}
\widetilde x_{i, k} = & A_{i,1,k}l(x_i) + A_{i,2,k}m(x_i) + A_{i,3,k}h(x_i),\\ 
\widetilde x_{j, k} = & \sum_{d=1}^{L_j} A_{j,d,k}l_d(x_j)   
\end{align}
\end{subequations}
respectively.

\subsubsection{The second layer}
The second layer of the rule module calculates rule firing strength by a connection matrix $\mathbf{M}$ of dimension $H \times K$. The $k^{th}$ rule is constructed as a combination of $\widetilde x_{1, k},\ldots, \widetilde x_{H, k}$ from the previous layer. An entry $M_{i,k}$ in the connection matrix $\mathbf{M}$ denotes the contribution of $x_i$ to the $k^{th}$ rule. Entries in the connection matrix are all trainable and constrained to $[0, 1]$ the hyperbolic tangent activation function, and a higher value indicates a higher contribution. In this layer, we define a parametrized T-norm to calculate $r_k$, the firing strength of the $k^{th}$ rule. 

With $0< \epsilon_2<1$, let $g_{\epsilon_2}: [0, \infty) \to [0, \infty)$ and its inverse function $g_{\epsilon_2}^{-1}$ be defined as
\begin{subequations}
\begin{align}
g_{\epsilon_2}\left(x\right) = &\frac{\epsilon_2}{1-\epsilon_2}\left(1-x^{\frac{\epsilon_2-1}{\epsilon_2}}\right),\\
g_{\epsilon_2}^{-1}\left(z\right) = &\left(1-\frac{1-\epsilon_2}{\epsilon_2}z\right)^{\frac{\epsilon_2}{\epsilon_2-1}}. 
\end{align}
\label{eq:rule_1}
\end{subequations}

The parametrized T-norm on two inputs is defined as
\begin{equation}
\begin{split}
    T_{\epsilon_2}\left(x,y\right) = &g_{\epsilon_2}^{-1}(g_{\epsilon_2}\left(x\right) + g_{\epsilon_2}\left(y\right)) \\
        = & \left(x^{\frac{\epsilon_2-1}{\epsilon_2}} + y^{\frac{\epsilon_2-1}{\epsilon_2}} -1\right)^{\frac{\epsilon_2}{\epsilon_2-1}},
\end{split}
\label{eq:T-norm}
\end{equation}
which has the following asymptotic behavior:
\begin{subequations}
\begin{align}
\lim_{\epsilon_2 \to 1} T_{\epsilon_2}\left(x,y\right)&=xy, \\
\lim_{\epsilon_2 \to 0} T_{\epsilon_2}\left(x,y\right)&= \min(x,y), 
\end{align}
\end{subequations}
which means that the defined T-norm can be modulated between $\mathrm{product}$ and $\min$ by $\epsilon_2$.

Using this definition of the T-norm, $r_k$ is calculated by applying the T-norm to multiple inputs:
\begin{equation}
\begin{split}
    r_k = & T_{\epsilon_2}\left(\widetilde x_{1, k}^{M_{1,k}}, \widetilde x_{2, k}^{M_{2,k}}, \ldots, \widetilde x_{H, k}^{M_{H,k}}\right) \\
        = & g_{\epsilon_2}^{-1}\left(\sum_{i=1}^{H}g_{\epsilon_2}(\widetilde x_{i, k}^{M_{i,k}})\right) \\
        = &\left(\sum_{i=1}^{H} \widetilde x_{i, k}^{M_{i,k}\cdot\frac{\epsilon_2-1}{\epsilon_2}}-H+1\right)^{\frac{\epsilon_2}{\epsilon_2-1}}.
\end{split}
\label{eq:r_k}
\end{equation}

In Equation (\ref{eq:r_k}), entries in the connection matrix $\mathbf{M}$ are used as exponents. Taking the example of $\widetilde x_{1, k}^{M_{1,k}}$, a lower $M_{1,k}$ (closer to $0$) means $\widetilde x_{1, k}^{M_{1,k}}$ is closer to $1$, consequently it contributes less to $r_k$ with the proposed T-norm. Thus, a lower value in $\mathbf{M}$ indicates a lower contribution to the rule firing strength, and vice versa.

\subsection{Inference module}
Let $C$ denote the number of classes in the classification task. The inference layer has $C$ nodes, one for each class, that are fully connected to the rule layer nodes. The firing strength of each node $o_c$ is calculated using the rule firing strengths with an inference matrix $\mathbf{W}$ of dimension $K \times C$. An entry $W_{j, c}$ denotes the contribution of the $k^{th}$ rule to the $c^{th}$ class. Entries in the inference matrix are all trainable and positive. A higher value indicates a higher contribution. In this layer, we define a parametrized T-conorm to calculate $o_c$. 

The parametrized T-conorm on two inputs is written as
\begin{equation}
    Q_{\epsilon_3} \left(x, y\right) = \left(x^{\frac{1}{\epsilon_3}} +  y^{\frac{1}{\epsilon_3}}\right)^{\epsilon_3},
\end{equation}
where $0<\epsilon_3<1$. This T-conorm has the following asymptotic behavior:
\begin{subequations}
\begin{align}
\lim_{\epsilon_3 \to 1} Q_{\epsilon_3}\left(x,y\right)&=x+y, \\
\lim_{\epsilon_3 \to 0} Q_{\epsilon_3}\left(x,y\right)&= \max\left(x,y\right),
\end{align}
\end{subequations}
which means that the defined T-conorm can be modulated between $\mathrm{addition}$ and $\max$ by $\epsilon_3$.

Using this definition of the T-conorm, $o_c$ is calculated by applying the T-conorm to multiple inputs:
\begin{equation}
\begin{split}
    o_c = & Q_{\epsilon_3} \left(W_{1, c}r_1, W_{2, c}r_2,\ldots, W_{K, c}r_K\right) \\
        = & \left(\sum_{k=1}^{K} (W_{k, c}r_k)^{\frac{1}{\epsilon_3}}\right)^{\epsilon_3}.
\end{split}
\label{eq:o_c}
\end{equation}

After the calculation of $o_1, o_2, \ldots, o_C$, a softmax activation function is applied to generate probabilities $p_1, p_2, \ldots, p_C$ of being in each class, which are all in $[0, 1]$ with $\sum_{c=1}^{C}p_c = 1$. 

As $\sum_{c=1}^{C}p_c = 1$, we can set the number of “valid” nodes in the inference module to $C-1$ to avoid ambiguity in rule representation. For example, when performing binary classification $W_{:,0}$ can be set to 0 so that the model will only learn subspaces related to the positive class.

\subsection{Network Interpretation}
\label{sec:interpretation}
The proposed network can both extract rules and inject rules in a way that humans can understand. The entries in the attention matrix $\mathbf{A}$ and connection matrix $\mathbf{M}$ represent the contribution of individual concepts and individual variables to each rule. The entries in the inference matrix $\mathbf{W}$ gives the contribution of individual rules to each class. 

With $\mathbf{A}$ and $\mathbf{M}$, a contribution matrix $\mathbf{S}$ can be constructed that expresses the contribution of individual concepts to each rule in the model. The matrix $S$ is of the same dimension as attention matrix $\mathbf{A}$, i.e., it is a partition matrix formed by concatenating submatrices $\mathbf{S}_1, \mathbf{S}_2, \ldots, \mathbf{S}_H$. For an ordinal/continuous variable $x_i$, the corresponding submatrix $\mathbf{S}_i$ has dimension $3\times K$ and for a categorical variable $x_j$, $\mathbf{S}_j$ has dimension $L_j\times K$. The entries $S_{i,d, k}$ of $\mathbf{S}_i$ and $S_{j,d, k}$ of $\mathbf{S}_j$ are calculated as
\begin{subequations}
\begin{align}
    S_{i,d, k} = & A_{i,d,k} \times M_{i,k}, \hspace{3mm} d\in\{1,2,3\},\\
    S_{j,d, k} = & A_{j,d,k} \times M_{j,k}, \hspace{3mm} d\in\{1, \ldots, L_j\},
\end{align}
\end{subequations}
respectively, where $ k\in\{1,\ldots,K\}$.

The entry $S_{i,d,k}$ is the contribution of the $d^{th}$ concept of $x_i$ to the $k^{th}$ rule. $\mathbf{S}_{:,:,k}$ encodes the construction of the $k^{th}$ rule, while $W_{k,:}$ captures the relationship between classes and the $k^{th}$.

The following is a toy example further demonstrating how humanly understandable rules are represented in the network.

Given a dataset with four continuous input variable $x_1, x_2, x_3, x_4$ and a binary response (negative/positive), $\mathbf{A}, \mathbf{M}, \mathbf{W}$ are trained and $\mathbf{S}$ can be calculated. Let us assume that in the contribution matrix $\mathbf{S}$, $S_{1,1,1}, S_{2,3,1}, S_{2,2,2}$, and $S_{3,1,2}$ are close to $1$, with all other entries close to $0$. In the inference matrix $\mathbf{W}$, $W_{1,2}$ and $W_{2,2}$ are close to $1$ while $W_{1,1}$ and $W_{2,1}$ are close to $0$. From the given $\mathbf{S}$ and $\mathbf{W}$, we can summarize two rules from the trained network as follows:

\begin{itemize}
\item \textbf{IF} $x_1$ is low and $x_2$ is high, \textbf{THEN} the sample is positive;
\item \textbf{IF} $x_2$ is medium and $x_3$ is low, \textbf{THEN} the sample is positive.
\end{itemize}

The above two rules are represented in ($\mathbf{S}_{:,:,1}, W_{1,:}$) and ($\mathbf{S}_{:,:,2}, W_{2,:}$), respectively. The definitions of “low”, “medium” and “high” concepts can be extracted from the parameters in the encoding module. The extracted rules mimic human logic. They can be used to justify the network's decisions and contribute to knowledge discovery. 

In practice, the trained model may have some redundant rules. In this study, the correlation between each pair of rules are calculated. Rules with high correlation and concepts with less contributions will be removed for rule visualization. 

\subsection{Model training and network initialization}

The proposed network is trained by back-propagation with an Adam optimizer. A regular cross-entropy loss $loss_{cs}$ is calculated to train the classification model. Additionally, an $\ell_1$ norm-based regularization term $loss_{\ell_1}$ is added to the loss function to favor rules with a smaller number of concepts, which are more feasible to use in practice. In addition, the correlation among encoded rules is calculated as a loss term $loss_{corr}$ to avoid extracting redundant rules. The loss function can be written as:

\begin{subequations}
\begin{align}
loss_{total}=& loss_{ce} + \lambda_1 loss_{\ell_1} + \lambda_2 loss_{corr},\\
loss_{l1} =& \left\| vec(\mathbf{A}) \right\|_1 + \left\| vec(\mathbf{M}) \right\|_1 ,\\
loss_{corr} =& \sum_{i=1}^{H-1}\sum_{j=i+1}^{H} vec(\mathbf{S}_{:,:,i})vec(\mathbf{S}_{:,:,j})
\end{align}
\end{subequations}
where $\lambda_1$ and $\lambda_2$ control the magnitude of the $\ell_1$ norm-based regularization term and correlation based regularization term, respectively. $vec(\cdot)$ denotes the vectorization of a matrix.

In this study, for simplicity, $\epsilon_1, \epsilon_2, \epsilon_3$ are constrained to be equal. They are initialized as 0.99 at the beginning of training and are gradually reduced with the number of training steps. The scheduling of the $\epsilon$ values can be written as
\begin{equation}
    \epsilon = max(\epsilon_{min}, \epsilon\cdot \gamma^{training\_steps}),
\end{equation}
where $\gamma$ is the decay rate that can be tuned as a hyperparameter. From our preliminary analysis, $\gamma=0.99$ usually is a good choice. $\epsilon_{min}$ is another hyperparameter, whose optimal value varies with different applications. The hyperparameter tuning strategy will be discussed in the next section. Our experiments show that starting with $\epsilon=0.99$ and reducing $\epsilon$ improves model optimization (as discussed in Section \ref{sec:exp_1}). 

Before model training, trainable parameters will be randomly initialized. To improve performance, especially when the size of the training dataset is small, practical rules from domain knowledge can be used to initialize the network. Revisiting the toy example in Section \ref{sec:interpretation}, if the extracted rules were instead previously known within the application domain, the matrices $\mathbf{A}, \mathbf{M},$ and $\mathbf{W}$ in the network could then be initialized as:

\begin{itemize}
\item $\mathbf{A}$: $A_{1,1,1}, A_{2,3,1}, A_{2,2,2}, A_{3,1,2}$ have a higher value and other entries in $A_{:,:,1}$ and $A_{:,:,2}$ have a lower value;
\item $\mathbf{M}$: $M_{1,1}, M_{2,1}, M_{2,2}, M_{3,2}$ have a higher value and other entries in $M_{:,1}$ and $M_{:,2}$ have a lower value;
\item $\mathbf{W}$: $W_{1,2}, W_{2,2}$ have a high value and $W_{1,1}, W_{2,1}$ have a low value;
\item Other entries in $\mathbf{A}$, $\mathbf{M}$, and $\mathbf{W}$ are randomly initialized.
\end{itemize}

\section{Datasets and Experimental Settings}
\subsection{Synthetic datasets}
Two synthetic datasets were built by simulating features with fixed distributions and rules to generate responses. The ground truth rules from the synthetic datasets can be used to assess a method's capability in extracting humanly understandable knowledge from the data and modeling the relationship between inputs and responses. In addition, with ground truth rules, synthetic datasets can be used to assess whether the proposed method can benefit from existing knowledge.

For each dataset, a 10-fold cross-validation was used for performance evaluation. In each iteration, the dataset was randomly split into the training set (64\%), validation set (16\%), and test set (20\%).

\subsubsection{Synthetic dataset 1}
\label{sec:synthetic_dataset_1}
Eight input variables were simulated as: $x_1 \sim \mathcal{N}(0,\,2)$, $x_2 \sim \mathcal{N}(5,\,3)$, $x_3 \sim \mathcal{N}(-1,\,5)$, $x_4 \sim \mathcal{N}(1,\,2)$, $x_5 \sim \mathcal{N}(-2,\,1)$, $x_6 \sim \mathrm{Bernoulli}(0.5)$, $x_7 \sim \mathcal{N}(0,\,1)$, $x_8 \sim \mathcal{N}(0,\,1)$. If any of the following rules apply to one observation, then this observation is positive and otherwise negative:

\begin{itemize}
\item Rule A: $x_2<3.8$ and $x_3>-2$ and $x_6=1$;
\item Rule B: $x_2>6.3$ and $x_3>-2$ and $x_6=1$;
\item Rule C: $x_1<1$ and $x_4>2$ and $x_6=0$;
\item Rule D: $x_3>0$ and $x_5>-1$ and $x_6=0$;
\item Rule E: $x_1<1$ and $x_5>-1.5$ and $x_6=0$.
\end{itemize}

Additionally, random noise sampled from $\mathcal{N}(0,\,0.01)$ are added to input variables. From the above rules we can readily observe that the response of one observation doesn't rely on $x_7$ and $x_8$. $x_7$ and $x_8$ are used as irrelevant variables to assess the model's resilience to redundant features.

\subsubsection{Synthetic dataset 2}
Nine input variables were simulated as: $x_1 \sim \mathcal{N}(0,\,2)$, $x_2 \sim \mathcal{N}(5,\,3)$, $x_3 \sim \mathcal{N}(-1,\,5)$, $x_4 \sim \mathcal{N}(1,\,2)$, $x_5 \sim \mathcal{N}(-2,\,1)$, $x_6 \sim \mathcal{N}(-1,\,4.4)$, $x_7 \sim \mathcal{N}(0,\,1.2)$, $x_8 \sim \mathcal{N}(0,\,1)$, $x_9 \sim \mathcal{N}(0,\,1)$. The sample is positive if $(x_1 + 0.5x_2+x_3)^2/(1+e^{x_6}+2x_7)<1$.

Unlike synthetic dataset 1, which is built from rules, a highly non-linear function is used to assign the response. Though such a relationship between input variables and responses rarely exists for clinical applications, this dataset is used to determine if the proposed network can still have achieve good performance by approximating the complicated relation as simple rules.

\subsection{Heart failure dataset}
A HF dataset is created to train a classification model that identifies patients eligible for advanced therapies. For this analysis, we focused our analysis on the timing of LVAD implantation and urgent HT as these urgent transplants occur on the order of months and can be predicted based on the time of transplant listing. Two cohorts were used in this study. 

\subsubsection{REVIVAL cohort} The REVIVAL (Registry Evaluation of Vital Information for VADs in Ambulatory Life) registry contains information on 400 patients with advanced systolic HF from 21 US medical centers. As part of the registry, patients were evaluated at up to 6 pre-specified time points over a 2-year period and underwent relevant examinations. At each time point, investigators were asked to record whether the participant had been evaluated for HT or LVAD and the result of that evaluation. Death, HT, and durable MCS implantation were study endpoints with no additional follow-up. For purposes of this analysis, study participants were labeled at each time point as appropriate (positive) or not appropriate (negative) for advanced therapies. In total, the cohort contains 96 positive samples from 62 patients, and 1336 negative samples from 339 patients.

\begin{table}[b]
\caption{Ratio of patients from different groups in training, validation, and test sets in one iteration.}
\label{tab:heart-failure-split}
\begin{tabular}{@{}cccc@{}}
\toprule
                                                                                            & Training set & Validation set & Test set \\ \midrule
\begin{tabular}[c]{@{}c@{}}Patients in REVIVAL \\ with advanced therapy (n=64)\end{tabular} & 0\%            & 50\%           & 50\%     \\ \midrule
\begin{tabular}[c]{@{}c@{}}Patients in REVIVAL \\ w/o advanced therapy (n=339)\end{tabular} & 80\%         & 10\%           & 10\%     \\ \midrule
\begin{tabular}[c]{@{}c@{}}Patients in \\ INTERMACS (n=2998)\end{tabular}                   & 100\%        & 0\%           & 0\%        \\ \bottomrule
\end{tabular}
\end{table}

\begin{table}[b]
\caption{Performance of the proposed model on synthetic dataset 1 with $N=400$ for different $\epsilon$ settings using 10-fold cross-validation. For the first four rows, $\epsilon$ was initialized to 0.99 and was gradually reduced to $\epsilon_{min}$ during training. For the last four rows, the value of $\epsilon$ was fixed during the training process.}
\label{tbl: performance_exp1}
\centering
\begin{tabular}{@{}cccccc@{}}
\toprule
Model & Accuracy & Recall & Precision & F1 & AUC \\ \midrule
$\epsilon_{min}$ = 0.8 & \begin{tabular}[c]{@{}c@{}}0.955 \\ (0.025)\end{tabular} & \begin{tabular}[c]{@{}c@{}}0.911 \\ (0.073)\end{tabular} & \begin{tabular}[c]{@{}c@{}}0.955 \\ (0.038)\end{tabular} & \begin{tabular}[c]{@{}c@{}}0.883 \\ (0.040)\end{tabular} & \begin{tabular}[c]{@{}c@{}}0.986 \\ (0.016)\end{tabular} \\ \hline
$\epsilon_{min}$ = 0.4 & \begin{tabular}[c]{@{}c@{}}0.959 \\ (0.030)\end{tabular} & \begin{tabular}[c]{@{}c@{}}0.904 \\ (0.073)\end{tabular} & \textbf{\begin{tabular}[c]{@{}c@{}}0.972\\  (0.035)\end{tabular} }& \begin{tabular}[c]{@{}c@{}}0.888 \\ (0.048)\end{tabular} & \begin{tabular}[c]{@{}c@{}}0.991 \\ (0.010)\end{tabular} \\ \hline
$\epsilon_{min}$ = 0.2 & \begin{tabular}[c]{@{}c@{}}0.961\\  (0.026)\end{tabular} &\textbf{ \begin{tabular}[c]{@{}c@{}}0.919 \\ (0.087)\end{tabular}} & \begin{tabular}[c]{@{}c@{}}0.968 \\ (0.039)\end{tabular} & \textbf{\begin{tabular}[c]{@{}c@{}}0.892 \\ (0.045)\end{tabular} }& \textbf{\begin{tabular}[c]{@{}c@{}}0.992 \\ (0.008)\end{tabular} }\\ \hline
$\epsilon_{min}$ = 0.1 & \begin{tabular}[c]{@{}c@{}}0.901 \\ (0.053)\end{tabular} & \begin{tabular}[c]{@{}c@{}}0.856 \\ (0.146)\end{tabular} & \begin{tabular}[c]{@{}c@{}}0.865 \\ (0.089)\end{tabular} & \begin{tabular}[c]{@{}c@{}}0.803 \\ (0.093)\end{tabular} & \begin{tabular}[c]{@{}c@{}}0.949\\  (0.056)\end{tabular} \\ \hline
Fixed $\epsilon$ = 0.8 &\textbf{ \begin{tabular}[c]{@{}c@{}}0.966 \\ (0.023)\end{tabular} }& \begin{tabular}[c]{@{}c@{}}0.903 \\ (0.083)\end{tabular} & \begin{tabular}[c]{@{}c@{}}0.964 \\ (0.019)\end{tabular} & \begin{tabular}[c]{@{}c@{}}0.886 \\ (0.037)\end{tabular} & \begin{tabular}[c]{@{}c@{}}0.978 \\ (0.019)\end{tabular} \\ \hline
Fixed $\epsilon$ = 0.4 & \begin{tabular}[c]{@{}c@{}}0.939 \\ (0.040)\end{tabular} & \begin{tabular}[c]{@{}c@{}}0.867 \\ (0.086)\end{tabular} & \begin{tabular}[c]{@{}c@{}}0.948 \\ (0.056)\end{tabular} & \begin{tabular}[c]{@{}c@{}}0.857 \\ (0.064)\end{tabular} & \begin{tabular}[c]{@{}c@{}}0.964 \\ (0.024)\end{tabular} \\ \hline
Fixed $\epsilon$ = 0.2 & \begin{tabular}[c]{@{}c@{}}0.786 \\ (0.041)\end{tabular} & \begin{tabular}[c]{@{}c@{}}0.519 \\ (0.190)\end{tabular} & \begin{tabular}[c]{@{}c@{}}0.803 \\ (0.109)\end{tabular} & \begin{tabular}[c]{@{}c@{}}0.558 \\ (0.132)\end{tabular} & \begin{tabular}[c]{@{}c@{}}0.819 \\ (0.117)\end{tabular} \\ \hline
Fixed $\epsilon$ = 0.1 & \begin{tabular}[c]{@{}c@{}}0.789 \\ (0.062)\end{tabular} & \begin{tabular}[c]{@{}c@{}}0.552 \\ (0.237)\end{tabular} & \begin{tabular}[c]{@{}c@{}}0.689 \\ (0.255)\end{tabular} & \begin{tabular}[c]{@{}c@{}}0.560 \\ (0.216)\end{tabular} & \begin{tabular}[c]{@{}c@{}}0.855 \\ (0.081)\end{tabular} \\ \bottomrule
\end{tabular}
\end{table}

\begin{table*}[]
\caption{Performance of ML methods on synthetic dataset 1 with $N=400$ using 10-fold cross-validation.}
\label{tbl: performance_exp1_compare}
\centering
\begin{tabular}{@{}cccccccc@{}}
\toprule
Model & Accuracy      & Recall        & Precision     & F1            & AUC                  & Transparent \\ \midrule
Proposed                       & 0.960 (0.023) & 0.933 (0.054) & 0.953 (0.060) & 0.893 (0.032) & 0.994 (0.005) &  Yes \\ \midrule
EBM                        & 0.835 (0.027) & 0.678 (0.060) & 0.807 (0.060) & 0.688 (0.045) & 0.924 (0.018) &  Yes \\
Logistic Regression        & 0.724 (0.029) & 0.344 (0.078) & 0.692 (0.098) & 0.413 (0.070) & 0.701 (0.065) &  Yes \\
Naïve Bayes                & 0.734 (0.032) & 0.363 (0.089) & 0.721 (0.114) & 0.434 (0.082) & 0.803 (0.035) &  Yes \\
Decision Tree              & 0.933 (0.046) & 0.907 (0.056) & 0.901 (0.090) & 0.855 (0.064) & 0.938 (0.040) &  Yes \\
Fuzzy Inference Classifier & 0.680 (0.036) & 0.456 (0.102) & 0.540 (0.076) & 0.441 (0.071) & 0.668 (0.056) &  Yes \\ \midrule
Random Forest              & 0.924 (0.015) & 0.826 (0.062) & 0.944 (0.037) & 0.832 (0.028) & 0.981 (0.006) &  No  \\
XGBoost                    & 0\textbf{.977 (0.013)} & \textbf{0.959 (0.031)} & \textbf{0.975 (0.028)} & \textbf{0.919 (0.020)} & \textbf{0.996 (0.003) }& No  \\
SVM                        & 0.821 (0.038) & 0.641 (0.076) & 0.796 (0.077) & 0.661 (0.061) & 0.897 (0.026) &  No \\ \bottomrule
\end{tabular}
\end{table*}

\subsubsection{INTERMACS cohort} The INTERMACS (Interagency Registry for Mechanically Assisted Circulatory Support) registry is a North American registry of adults who received an FDA approved durable MCS device for the management of advanced HF. The registry includes clinical data on all adults $\geq$ 19 years of age who received a device at one of 170 active INTERMACS centers. The registry includes information on patient demographics, clinical data before and at the time of MCS implantation, and clinical outcomes up to one year post-MCS implantation or until HT. For this analysis, data was extracted at the time of LVAD implantation and patients classified as “appropriate for advanced therapies.” In total, the cohort contains 7781 positive samples from 7781 patients.

Patients from the two cohorts were combined to form a larger dataset. 23 clinical variables were selected by clinicians and used in this study including heart rate, systolic blood pressure (SYSBP), sodium concentration, albumin concentration, uric acid concentration, total distance walked in 6 minutes (DISTWLK), gait speed during a 15 feet walk test, left ventricular dimension in diastole (LVDEM), left ventricular ejection fraction (EF), eight-item Patient Health Questionnaire depression scale (PHQ-8) score score, mitral regurgitation (MITRGRG), lymphocyte percentage (LYMPH), total cholesterol (TCH), hemoglobin (HGB), age, sex, comorbidity index, glomerular filtration rate (GFR), pulse pressure, treatment with cardiac resynchronization therapy (AR), need for temporary MCS device, treatment with guideline directed medical therapy (GDMT) for heart failure, and peak oxygen consumption during a maximal cardiopulmonary exercise test (pVO2). Note, in this study, EF denotes the ejection fraction severity score, that means a patient with a low ejection fraction has a high EF value.

Patient-wise splitting was performed to construct training, validation, and test sets, the details of which are shown in Table \ref{tab:heart-failure-split}. To facilitate model training, 5 approximate rules denoting eligibility for advanced therapies were collected from heart failure and transplant cardiologists:
\begin{itemize}
\item Rule A: EF is high, and pVO2 is low;
\item Rule B: EF is high, and DISTWLK is low;
\item Rule C: Age is high, EF is high, and SYSBP is low;
\item Rule D: EF is high, and MITRGRG is high;
\item Rule E: EF is high, and the GDMT is low;
\end{itemize}

\subsection{Experimental settings}
For synthetic datasets, a 10-fold cross-validation was used to evaluate model performance; and for heart failure dataset, the proposed data split in Table \ref{tab:heart-failure-split} was randomly repeated for 10 times to evaluate the model. A random search algorithm was applied using the training set and validation set for hyperparameter tuning, including learning rate, batch size, $\lambda_1$, $\lambda_2$, and $\epsilon_{min}$. The model trained with the optimal combinations of hyperparameters was then evaluated on the test set. The performance of the proposed network will be presented as the average and standard deviation ($\mathrm{std}$) from 10 iterations.

For comparison, several popular “black box” machine learning algorithms were chosen, including random forest, SVM, and XGBoost. In addition, several interpretable models were chosen including logistic regression, decision tree, and Explainable Boosting Machine (EBM, a type of GAM) \cite{nori2019interpretml}, and a fuzzy inference classifier \cite{meher2007new}. Those models have the same hyper-parameter tuning and model evaluation as the proposed algorithm. Detailed implementation information for these models is described in Appendix \ref{sec:a1}.

Accuracy, recall, precision, F1, and area under the ROC curve (AUC) were calculated to evaluate the performance of the trained classifiers.

\section{Results and Discussion}

\begin{figure*}[] 
\centering
\scalebox{0.98}{
\includegraphics[width=7.0in]{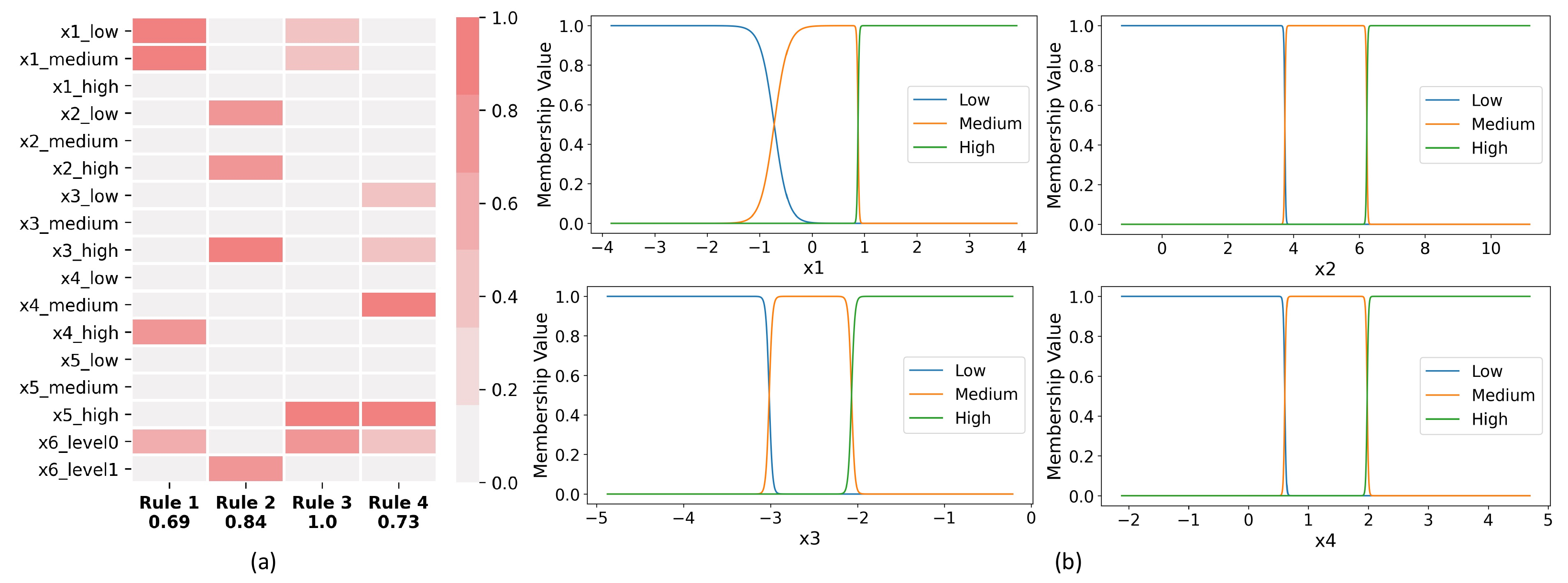}
}
\caption{Interpretation of a trained model on synthetic dataset 1 with $N=400$. (a) Visualization of four rules contributing to the positive class, which are summarized from the trained model. Rules are visualized in individual columns with each row correspond to concept. For example, “x1\_low” means “the value of $x_1$ is low”. The contribution of individual concepts to individual rules are shown in color; (b) Membership functions for “low”, “medium”, and “high” concepts of $x_1, x_2, x_3$, and  $x_4$ in the encoding module, respectively.}
\label{fig:rules_exp_1_1}
\end{figure*}

\subsection{Synthetic dataset 1 ($N=400$)}
\label{sec:exp_1}
Let $N$ denote the number of observations in a given dataset. Several experiments were performed with differently sized simulated datasets. In this section, we discuss the performance of the proposed method on synthetic dataset 1 when $N=400$.

The first experiment starts with $N=400$.The proposed network was trained using 80\% of the data and tested on 20\% of the data. The percentage of positive samples is 34.25\%, and the percentages of samples with Rule A, Rule B, Rule C, Rule D, Rule E are 8.25\%, 7.50\%, 9.00\%, 2.00\%, and 10.75\%, respectively.  

Table \ref{tbl: performance_exp1} depicts the performance of the proposed algorithm with different $\epsilon_{min}$ on the test sets from a 10-fold cross-validation. We can observe that model training benefited from decreasing $\epsilon_{min}$ from 0.8 to 0.2, but the performance of the trained model decreased when  $\epsilon_{min}$ was decreased to 0.1. We also evaluated the model with a fixed $\epsilon$, rather than gradually decreasing it from 0.99. While fixing $\epsilon$ at 0.8 leads to comparable performance with the model using $\epsilon_{min}=0.8$, the performance of the models with a smaller fixed $\epsilon$ value decreased significantly. Our results show the effectiveness of the algorithm that gradually decreases $\epsilon$ during the training. Using this dataset, the proposed network with a reasonable degree of piecewise linearity has a better performance.

Table \ref{tbl: performance_exp1_compare} describes the performance of the proposed method where $\epsilon_{min}$ is tuned on the validation set in each iteration. The performance of the proposed network is compared with that of other machine learning algorithms. From Table \ref{tbl: performance_exp1_compare}, we can see that the proposed network achieved significantly better performance than other interpretable models and had comparable performance to the XGBoost model, which is the best among the other established machine learning algorithms. 

To examine the proposed network's ability to learn rules from the dataset, we summarized rules contributing to the positive class from a trained network. Those rules are visualized in Figure \ref{fig:rules_exp_1_1} (a). Comparing the learned rules with rules in Section \ref{sec:synthetic_dataset_1}, we can observe that Rule 1 corresponds to Rule C; Rule 2 corresponds to a union of Rule A and Rule B; Rule 3 corresponds to Rule E; and Rule 4 is closest to Rule D. Membership functions of the variables involved in Rule 1 and Rule 2 are visualized in Figure \ref{fig:rules_exp_1_1} (b) and we can observe a great match. For example, the membership value of $x_2$ to the “low” concept is high when $x_2$ smaller than 3.7 and the membership value of $x_2$ to the “high” concept is high when $x_2$ is larger than 6.2. Simple thresholds were used to construct synthetic dataset 1, and for this reason the fuzzy regions in the membership functions are very narrow. From the interpretation in Figure \ref{fig:rules_exp_1_1}, the trained model learned the majority of rules used to construct the dataset. Rule 4 is close to Rule D but with two additional concepts that are misidentified as related to the class. This may be due to only 2.00\% of samples in the dataset being consistent with Rule D, making it more challenging to learn from the data. In addition, from Figure \ref{fig:rules_exp_1_1} (a), concepts from $x_7$ and $x_8$ are not shown because their significance to learned rules is too low. This demonstrates that the proposed network can identify and exclude irrelevant variables.

\begin{table*}[]
\caption{Performance of ML methods on the synthetic dataset 1 with $N=50$ using 10-fold cross-validation.}
\label{tbl:performance_exp_2_n50}
\centering
\begin{tabular}{@{}cccccccc@{}}
\toprule
Model                      & Accuracy      & Recall        & Precision     & F1            & AUC           & Transparent \\ \midrule
Proposed (None) & 0.640 (0.143) & 0.550 (0.292) & 0.518 (0.249) & 0.473 (0.236) & 0.688 (0.213) & Yes \\
Proposed (Rule A) & 0.670 (0.110) & 0.575 (0.275) & 0.543 (0.238) & 0.504 (0.223) & 0.710 (0.188) & Yes \\
Proposed (Rule B) & 0.670 (0.135) & 0.600 (0.255) & 0.646 (0.211) & 0.535 (0.170) & 0.658 (0.183) & Yes \\
Proposed (Rule C) & 0.690 (0.104) & 0.625 (0.202) & 0.658 (0.197) & 0.566 (0.129) & 0.698 (0.158) & Yes \\
Proposed (Rule D) & 0.730 (0.142) & \textbf{0.675 (0.251) }& 0.658 (0.282) & \textbf{0.607 (0.225)} & 0.710 (0.194) & Yes \\
Proposed (Rule E) & 0.700 (0.190) & 0.600 (0.229) & 0.710 (0.259) & 0.573 (0.202) & \textbf{0.740 (0.191) }& Yes \\
Proposed (Rule F, partially correct) & 0.680 (0.183) & 0.600 (0.200) & 0.665 (0.278) & 0.565 (0.196) & 0.688 (0.206) & Yes \\
Proposed (Rule G, partially correct) & 0.700 (0.210) & 0.625 (0.280) & 0.605 (0.308) & 0.566 (0.276) & 0.652 (0.213) & Yes \\
Proposed (Rule H, partially correct) & \textbf{0.750 (0.112) }& 0.575 (0.195) & \textbf{0.775 (0.197) }& 0.593 (0.176) & \textbf{0.740 (0.152)} & Yes \\ \midrule
EBM                        & 0.650 (0.120) & 0.500 (0.224) & 0.562 (0.260) & 0.469 (0.192) & 0.670 (0.151) &  Yes         \\
Logistic Regression        & 0.610 (0.145) & 0.425 (0.275) & 0.512 (0.339) & 0.395 (0.236) & 0.583 (0.181) &  Yes         \\
Naïve Bayes                & 0.640 (0.120) & 0.475 (0.208) & 0.552 (0.159) & 0.457 (0.178) & 0.629 (0.174) &  Yes         \\
Decision Tree              & 0.530 (0.200) & 0.425 (0.317) & 0.398 (0.263) & 0.361 (0.261) & 0.527 (0.203) &  Yes         \\
Fuzzy Inference Classifier & 0.520 (0.117) & 0.525 (0.208) & 0.416 (0.120) & 0.413 (0.146) & 0.550 (0.103) &  Yes         \\ \midrule
Random Forest              & 0.650 (0.081) & 0.475 (0.236) & 0.580 (0.275) & 0.450 (0.176) & 0.619 (0.168) &  No          \\
XGBoost                    & 0.650 (0.186) & 0.600 (0.300) & 0.591 (0.275) & 0.521 (0.238) & 0.675 (0.187) &  No          \\
SVM                        & 0.580 (0.075) & 0.125 (0.230) & 0.250 (0.403) & 0.130 (0.204) & 0.521 (0.173) &  No          \\ \bottomrule
\end{tabular}
\end{table*}

\begin{figure*}[] 
\centering
\scalebox{0.98}{
\includegraphics[width=7.0in]{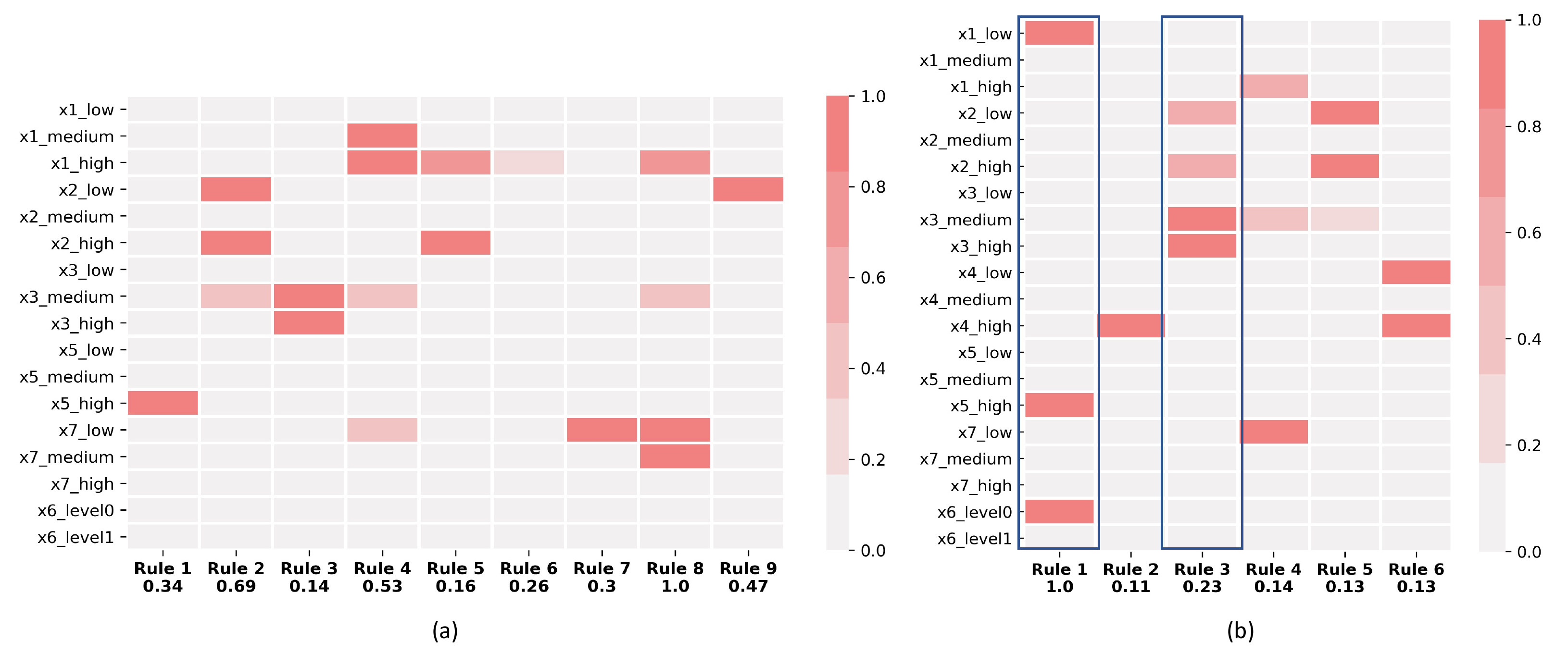}
}
\caption{Rules contributing to the positive class learned by the proposed network on synthetic dataset 1 with $N=50$. (a) Model parameters were randomly initialized; (b) $\mathbf{A}_{:,:,1}$, $M_{:,1}$, and $W_{1,:}$ were initialized by Rule $H$ while other entries were initialized with the same values in (a).}
\label{fig:rules_exp_1_2}
\end{figure*}

\subsection{Synthetic dataset 1 ($N=50$)} In the second experiment, we used synthetic dataset 1 with $N=50$. The percentage of positive samples is 42.00\%, and the percentages of samples with Rules A-E are  14.00\%,  14.00\%,  4.00\%,  4.00\%,  and 12.00\%,  respectively. In this experiment, we investigated the performance of the proposed network with a small training set and if initiating the network with existing knowledge would enable the model to learn more accurate rules.

\begin{table*}[]
\caption{Performance of ML methods on the synthetic dataset 2 with $N=400$ using 10-fold cross-validation.}
\label{tbl:exp_2}
\centering
\begin{tabular}{@{}cccccccc@{}}
\toprule
Model    & Accuracy      & Recall        & Precision     & F1            & AUC                   & Transparent \\ \midrule
Proposed & 0.714 (0.041) & 0.738 (0.067) & 0.693 (0.062) & 0.657 (0.045) & 0.801 (0.040)   & Yes         \\ \midrule
EBM      & 0.736 (0.028) & 0.686 (0.047) & 0.731 (0.044) & 0.660 (0.028) & 0.826 (0.042)  & Yes         \\
Logistic   Regression      & 0.746 (0.046) & 0.703 (0.084) & 0.738 (0.053) & 0.671 (0.058)  & 0.774 (0.073) & Yes \\ 
Naïve Bayes                & 0.723 (0.047) & 0.665 (0.078) & 0.720 (0.068) & 0.642 (0.054)  & 0.807 (0.051) & Yes \\
Decision Tree              & 0.674 (0.046) & 0.616 (0.069) & 0.660 (0.058) & 0.589 (0.052) & 0.679 (0.050)  & Yes \\
Fuzzy Inference Classifier & 0.654 (0.048) & 0.408 (0.090) & 0.721 (0.076) & 0.475 (0.084) & 0.761 (0.037)  & Yes \\\midrule
Random Forest              & 0.734 (0.040) & 0.692 (0.030) & 0.726 (0.058) & 0.660 (0.034) & 0.827 (0.035)  & No  \\
XGBoost                    & 0.734 (0.043) & 0.705 (0.072) & 0.714 (0.043) & 0.662 (0.054) & 0.837 (0.033)  & No  \\
SVM                        &\textbf{ 0.781 (0.074)} & \textbf{0.741 (0.077) }& \textbf{0.780 (0.094)} & \textbf{0.712 (0.079}) &\textbf{ 0.871 (0.066)} & No  \\ \bottomrule
\end{tabular}
\end{table*}

Table \ref{tbl:performance_exp_2_n50} has three blocks, presenting the performance of the proposed networks, established interpretable ML methods, and established black-box ML methods on synthetic dataset 1 ($N=50$), respectively. The first block shows the performance of the proposed network without and with existing knowledge. The performance of the proposed network with random initialization is shown in the first row of the first block, followed by the performance of the proposed network initialized with existing knowledge (rules). Rules A through E are fully correct as described in Section \ref{sec:synthetic_dataset_1} while Rules F through H are partially correct. In practical applications, it is very rare that the ground truth rule is available. As such, in this experiment, we only initialized $\mathbf{A}$, $\mathbf{M}$, and $\mathbf{W}$, while the parameters in the membership functions were randomly initialized. In addition, to investigate whether inexact domain knowledge can facilitate model training, we proposed the following three rules and assumed they lead to a positive class:
\begin{itemize}
\item Rule F: $x_2$ is “low” and $x_6=1$;
\item Rule G: $x_1$ is “low” and $x_5$ is “low” and $x_6=0$;
\item Rule H: $x_1$ is “low” and $x_5$ is “high” and $x_6=0$ and $x_7$ is “high”;
\end{itemize}
Rule F, G, and H are only partially correct. Compared with ground truth Rule A, the “high” concept of $x_3$ is missing in Rule F. In Rule G,  $x_5$ should be “high” rather than “low” as in Rule E. In Rule H, “high” concept of $x_7$ is actually irrelevant to the class.

From Table \ref{tbl:performance_exp_2_n50}, we first observe that because of the reduction in the size of the training set, performance decreased. Still, XGBoost achieves the best performance, and the proposed network with random initialization has a comparable performance to XGBoost. Second, we observe that the improvement can be achieved when the network was initialized with Rules A through E. Third, the model's performance increased when it was initialized with partially correct rules. This indicates that existing domain knowledge can help with model training even when the rules are vague and/or inexact.

In Figure \ref{fig:rules_exp_1_2}, we interpret and visualize the model trained from scratch and the model initialized with Rule H. From Figure \ref{fig:rules_exp_1_2} (a), we find that the learned rules are less accurate compared with Figure \ref{fig:rules_exp_1_1} (a) because of the reduced size of the training set. In Figure \ref{fig:rules_exp_1_2} (b), Rule 1 shows that even though the model was initialized with a partially correct rule, the model can identify that “high” $x_7$ doesn't contribute to the classification; and Rule 3 indicates that initializing the model with existing knowledge can also facilitate the model learning other rules.

\subsection{Synthetic dataset 2 ($N=400$)}
The responses in synthetic dataset 1 were constructed by rules, where a rule-based or tree-based machine learning algorithm may be more favorable. Therefore, responses in synthetic dataset 2 were built from a non-linear function to further explore the capacity of the proposed network in function approximation. A performance comparison of different ML models is presented in Table \ref{tbl:exp_2}. From the table, we can see that SVM achieved the best performance. The performance of the proposed network is lower than SVM but comparable with other machine learning algorithms.

\begin{figure}[h] 
\centering
\scalebox{0.98}{
\includegraphics[width=3.4in]{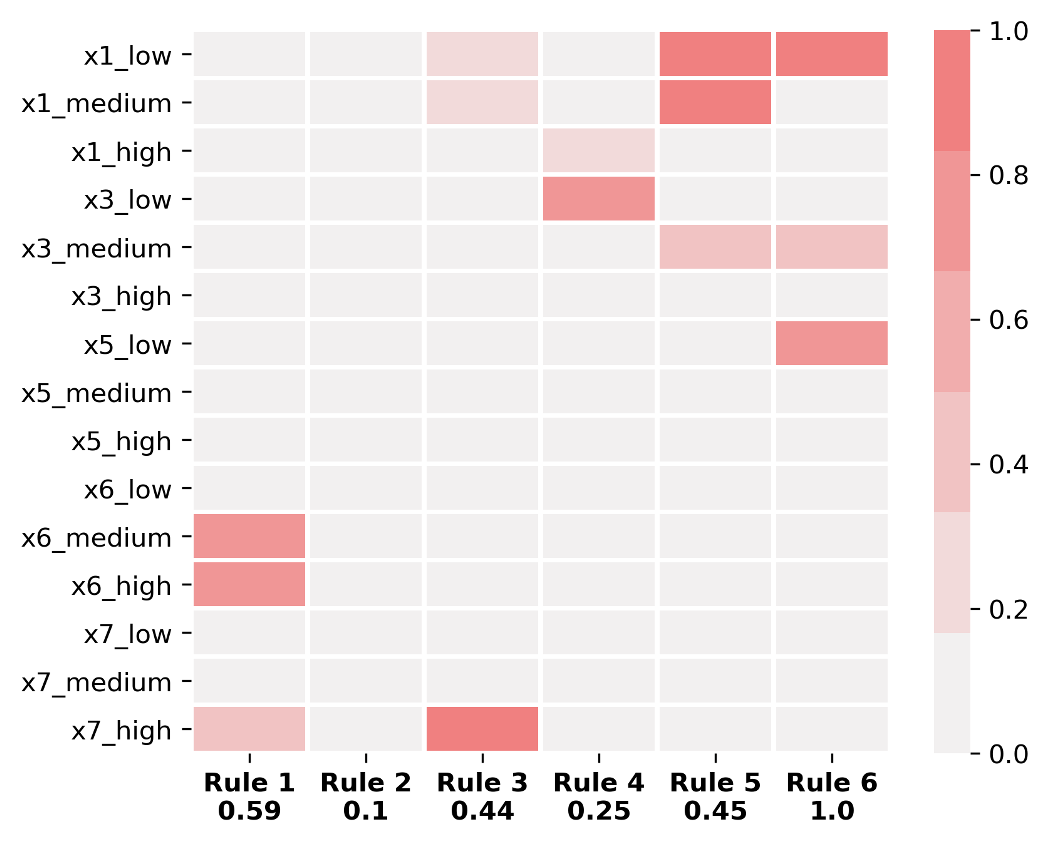}
}
\caption{Interpretation of a trained model on synthetic dataset 2 with $N=400$.}
\label{fig:rules_exp_2}
\end{figure}

\begin{table*}[]
\caption{Performance of ML methods on the test set of the heart failure dataset from 10 repetitions.}
\centering
\begin{tabular}{@{}ccccccc@{}}
\toprule
Model                          & Accuracy      & Recall        & Precision     & F1            & AUC           & Transparent \\ \midrule
Proposed (None)                & 0.735 (0.047) & 0.500 (0.069) & 0.384 (0.059) & 0.386 (0.047) & 0.730 (0.042) & Yes         \\
Proposed (with existing rules) & 0.718 (0.035) & \textbf{0.645 (0.125)} & 0.410 (0.045) & \textbf{0.452 (0.043)} & 0.753 (0.025) & Yes         \\ \midrule
EBM                            & 0.787 (0.018) & 0.122 (0.032) & 0.557 (0.150) & 0.173 (0.049) & 0.795 (0.034) & Yes         \\
Logistic Regression            & 0.783 (0.011) & 0.000 (0.000) & 0.000 (0.000) & 0.000 (0.000) & 0.541 (0.062) & Yes         \\
Naïve Bayes                    & 0.781 (0.012) & 0.012 (0.013) & 0.383 (0.435) & 0.019 (0.020) & 0.496 (0.025) & Yes         \\
Decision Tree                  & 0.787 (0.013) & 0.072 (0.043) & 0.600 (0.221) & 0.108 (0.061) & 0.593 (0.047) & Yes         \\
Fuzzy Inference Classifier     & 0.669 (0.182) & 0.422 (0.379) & 0.454 (0.170) & 0.262 (0.130) & 0.739 (0.048) & Yes         \\ \midrule
Random Forest                  & 0.782 (0.011) & 0.004 (0.012) & 0.029 (0.086) & 0.005 (0.016) &\textbf{ 0.834 (0.016)} & No          \\
XGBoost                        & \textbf{0.792 (0.013) }& 0.079 (0.035) &\textbf{ 0.659 (0.104)} & 0.123 (0.051) & 0.792 (0.029) & No          \\
SVM                            & 0.746 (0.037) & 0.116 (0.068) & 0.291 (0.181) & 0.130 (0.079) & 0.636 (0.069) & No          \\ \bottomrule
\label{tab:heat-failure}
\end{tabular}
\end{table*}

Rules extracted from the trained proposed network are presented in Figure \ref{fig:rules_exp_2}. We see that these rules capture meaningful information. Observations in this dataset were annotated as positive if $(x_1 + 0.5x_2+x_3)^2/(1+e^{x_6}+2x_7)<1$. Rule 1 shows that “high” levels of $x_6$ and $x_7$ lead to the positive class. In this dataset, $x_1$, $x_2$, and $x_3$ were simulated as: $x_1 \sim \mathcal{N}(0,\,2)$, $x_2 \sim \mathcal{N}(5,\,3)$, and $x_3 \sim \mathcal{N}(-1,\,5)$. As such, a “high” $x_1$ and "low" $x_3$ can lead $(x_1 + 0.5x_2+x_3)^2$ to a small value. A "low" or "medium"  $x_1$ and "medium" $x_3$ is another combination that can lead $(x_1 + 0.5x_2+x_3)^2$ to a small value. As expected, Rules 4 and 5 unite concepts from $x_1$ and $x_3$. From this analysis, we observe that the proposed network can learn simple rules in a format that humans can understand from a dataset that was constructed with a complicated non-linear function.

\subsection{Heart failure dataset}

\begin{figure}[] 
\centering
\scalebox{1}{
\includegraphics[width=3.2in]{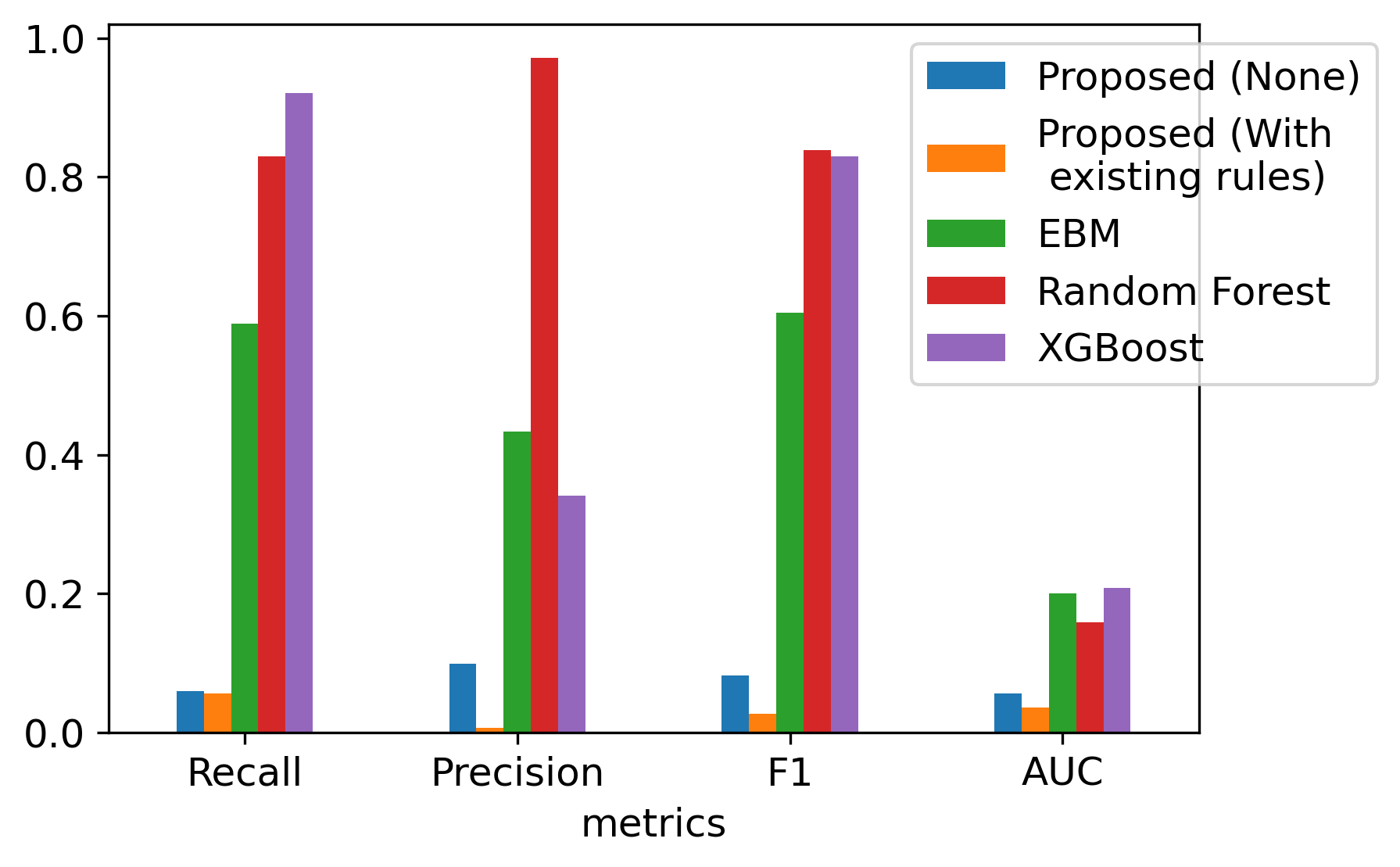}
}
\caption{Generalization error between the validation set and test set.}
\label{fig:diff}
\end{figure}

 We applied the proposed network to identify patients that are eligible for advanced therapies. From Table \ref{tab:heat-failure}, initializing the network with existing knowledge can greatly facilitate model performance. The proposed method had a lower AUC compared with EBM, Random Forest, and XGBoost. However, those models have low values in recall and F1-score, which means they tend to classify all samples as “negative”. In addition, those three methods achieved very high values on the validation set for all metrics, and this indicates severe overfitting on the validation set. Figure \ref{fig:diff} shows the generalization error between validation set and test set for five ML models. We can find the generalization errors for EBM, Random Forest, and XGBoost are very high. In contrast, the proposed method had a significantly smaller generalization error. Notably, integrating existing domain knowledge can not only improve the classification performance, but also further reduce the generalization error.
 
\begin{figure}[] 
\centering
\includegraphics[width=3.45in]{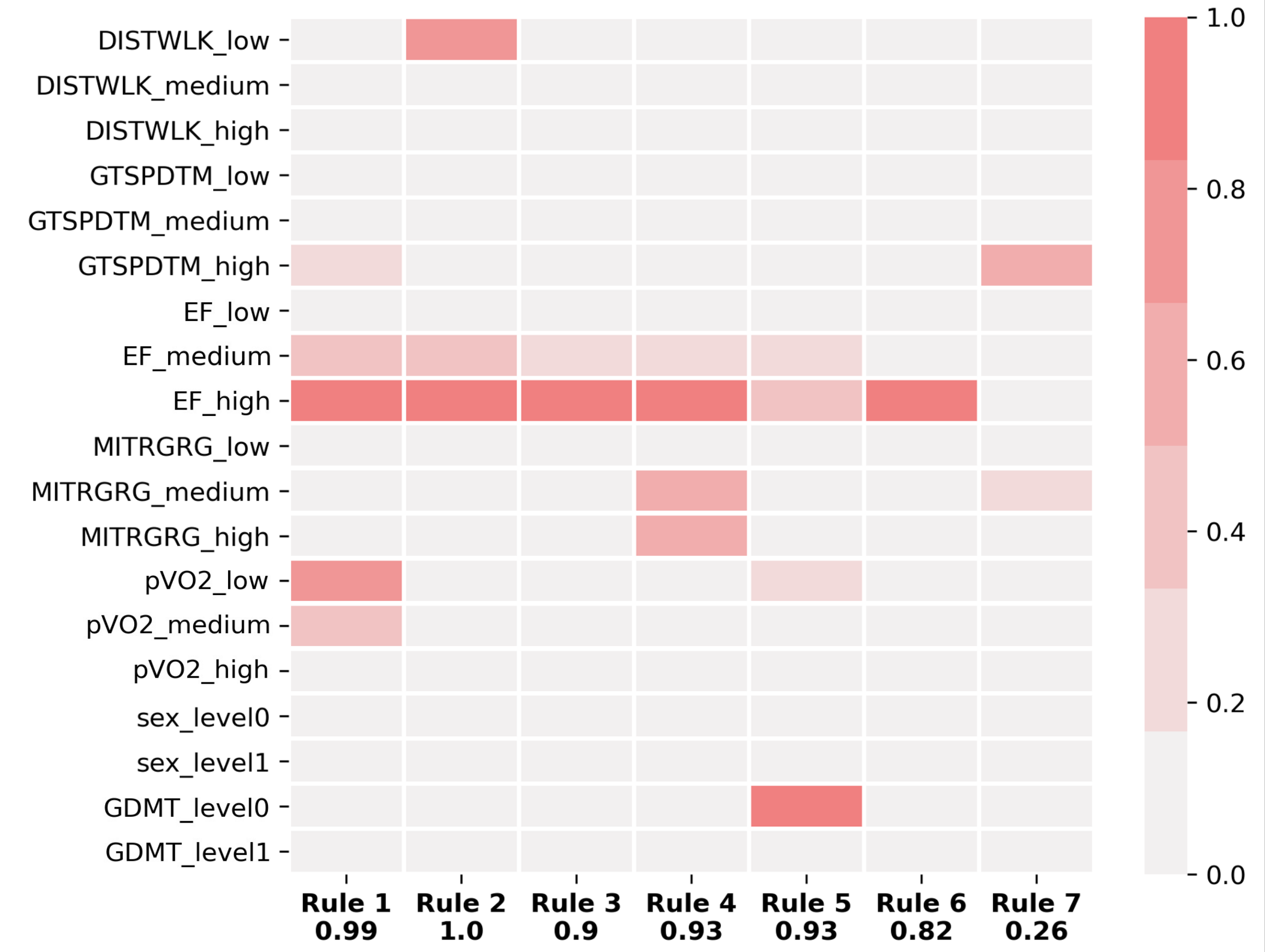}
\caption{Interpretation of a trained model on the heart failure dataset.}
\label{fig:rules_hf}
\end{figure}

 Figure \ref{fig:rules_hf} shows the learned rules of the trained model initialized with existing knowledge. These learned rules approximated those provided by heart failure cardiologists though in unique combinations and with additional learned features. All of the rules from heart failure cardiologists included a reduced ejection fraction and an objective marker of significant functional limitations, most often by cardiopulmonary exercise testing. As seen in Figure \ref{fig:rules_hf}, almost all rules learned by the model included ejection fraction as well as a second variable denoting a patient's functional tolerance, either by cardiopulmonary exercise testing, 6-minute walk distance, or by gait speed. Notably, while gait speed is an objective and valid measure of functional capacity, it was not included in any of the provided rules and thus represents learned knowledge.

\section{Conclusion}

In this study, we proposed a novel machine learning model that is transparent and interpretable. The proposed network was tested on both synthetic datasets and a real-world dataset. Our experimental results show that (1) the model can learn hidden rules from the dataset and represent them in a way that humans can understand; (2) the introduction of the smoothness factor enables the model to find the most suitable encoding functions and aggregation operators, which increases the performance of the proposed method; and (3) initializing the network with existing approximate domain knowledge can effectively improve model performance and generalizability, especially when the size of the training set is limited. Notably, the proposed network shows significantly improved generalizability when identifying patients with heart failure who would benefit from advanced therapies. The proposed algorithm is promising in building multiple other clinical (and non-clinical) decision-making applications.

The proposed network will be further extended and explored in future work. In the current optimization method, we use the same smoothness factor for encoding membership functions and aggregation operators. A simple linear decrease with the training steps was performed to optimize the smoothness factor. In future work, we will explore the possibility of optimizing the smoothness factors individually in respective modules with a more effective optimization method.


\section*{ACKNOWLEDGMENT}
The work was supported by the National Science Foundation under Grant No. 2014003.

\bibliographystyle{IEEEtran}
\bibliography{reference}

\begin{thebibliography}{10}
\providecommand{\url}[1]{#1}
\csname url@samestyle\endcsname
\providecommand{\newblock}{\relax}
\providecommand{\bibinfo}[2]{#2}
\providecommand{\BIBentrySTDinterwordspacing}{\spaceskip=0pt\relax}
\providecommand{\BIBentryALTinterwordstretchfactor}{4}
\providecommand{\BIBentryALTinterwordspacing}{\spaceskip=\fontdimen2\font plus
\BIBentryALTinterwordstretchfactor\fontdimen3\font minus
  \fontdimen4\font\relax}
\providecommand{\BIBforeignlanguage}[2]{{%
\expandafter\ifx\csname l@#1\endcsname\relax
\typeout{** WARNING: IEEEtran.bst: No hyphenation pattern has been}%
\typeout{** loaded for the language `#1'. Using the pattern for}%
\typeout{** the default language instead.}%
\else
\language=\csname l@#1\endcsname
\fi
#2}}
\providecommand{\BIBdecl}{\relax}
\BIBdecl

\bibitem{parikh2018heart}
K.~S. Parikh, K.~Sharma, M.~Fiuzat, H.~K. Surks, J.~T. George, N.~Honarpour,
  C.~Depre, P.~Desvigne-Nickens, R.~Nkulikiyinka, G.~D. Lewis \emph{et~al.},
  ``Heart failure with preserved ejection fraction expert panel report: current
  controversies and implications for clinical trials,'' \emph{JACC: Heart
  Failure}, vol.~6, no.~8, pp. 619--632, 2018.

\bibitem{benjamin137american}
E.~Benjamin, S.~Virani, C.~Callaway, A.~Chamberlain, A.~Chang, S.~Cheng,
  S.~Chiuve, M.~Cushman, F.~Delling, R.~Deo \emph{et~al.}, ``American heart
  association council on e, prevention statistics c, stroke statistics s (2018)
  heart disease and stroke statistics-2018 update: a report from the american
  heart association,'' \emph{Circulation}, vol. 137, no.~12, pp. e67--e492.

\bibitem{noorbakhsh2019artificial}
N.~Noorbakhsh-Sabet, R.~Zand, Y.~Zhang, and V.~Abedi, ``Artificial intelligence
  transforms the future of health care,'' \emph{The American journal of
  medicine}, vol. 132, no.~7, pp. 795--801, 2019.

\bibitem{caccomo2018fda}
S.~Caccomo, ``Fda permits marketing of artificial intelligence algorithm for
  aiding providers in detecting wrist fractures,'' \emph{FDA News Release},
  2018.

\bibitem{myszczynska2020applications}
M.~A. Myszczynska, P.~N. Ojamies, A.~M. Lacoste, D.~Neil, A.~Saffari, R.~Mead,
  G.~M. Hautbergue, J.~D. Holbrook, and L.~Ferraiuolo, ``Applications of
  machine learning to diagnosis and treatment of neurodegenerative diseases,''
  \emph{Nature Reviews Neurology}, vol.~16, no.~8, pp. 440--456, 2020.

\bibitem{senders2018machine}
J.~T. Senders, P.~C. Staples, A.~V. Karhade, M.~M. Zaki, W.~B. Gormley, M.~L.
  Broekman, T.~R. Smith, and O.~Arnaout, ``Machine learning and neurosurgical
  outcome prediction: a systematic review,'' \emph{World neurosurgery}, vol.
  109, pp. 476--486, 2018.

\bibitem{yao2020automated}
H.~Yao, C.~Williamson, J.~Gryak, and K.~Najarian, ``Automated hematoma
  segmentation and outcome prediction for patients with traumatic brain
  injury,'' \emph{Artificial Intelligence in Medicine}, vol. 107, p. 101910,
  2020.

\bibitem{davenport2019potential}
T.~Davenport and R.~Kalakota, ``The potential for artificial intelligence in
  healthcare,'' \emph{Future healthcare journal}, vol.~6, no.~2, p.~94, 2019.

\bibitem{zadeh1975fuzzy}
L.~A. Zadeh, ``Fuzzy logic and approximate reasoning,'' \emph{Synthese},
  vol.~30, no.~3, pp. 407--428, 1975.

\bibitem{takagi1985fuzzy}
T.~Takagi and M.~Sugeno, ``Fuzzy identification of systems and its applications
  to modeling and control,'' \emph{IEEE transactions on systems, man, and
  cybernetics}, no.~1, pp. 116--132, 1985.

\bibitem{chan2011diagnosis}
K.~Y. Chan, S.-H. Ling, T.~S. Dillon, and H.~T. Nguyen, ``Diagnosis of
  hypoglycemic episodes using a neural network based rule discovery system,''
  \emph{Expert Systems with Applications}, vol.~38, no.~8, pp. 9799--9808,
  2011.

\bibitem{zhang2017nonlinear}
R.~Zhang and J.~Tao, ``A nonlinear fuzzy neural network modeling approach using
  an improved genetic algorithm,'' \emph{IEEE Transactions on Industrial
  Electronics}, vol.~65, no.~7, pp. 5882--5892, 2017.

\bibitem{mikhalkin2006tropical}
G.~Mikhalkin, ``Tropical geometry and its applications,'' \emph{arXiv preprint
  math/0601041}, 2006.

\bibitem{doshi2017towards}
F.~Doshi-Velez and B.~Kim, ``Towards a rigorous science of interpretable
  machine learning,'' \emph{arXiv preprint arXiv:1702.08608}, 2017.

\bibitem{mittelstadt2019explaining}
B.~Mittelstadt, C.~Russell, and S.~Wachter, ``Explaining explanations in ai,''
  in \emph{Proceedings of the conference on fairness, accountability, and
  transparency}, 2019, pp. 279--288.

\bibitem{louppe2014understanding}
G.~Louppe, ``Understanding random forests: From theory to practice,''
  \emph{arXiv preprint arXiv:1407.7502}, 2014.

\bibitem{hohman2019s}
F.~Hohman, H.~Park, C.~Robinson, and D.~H.~P. Chau, ``S ummit: Scaling deep
  learning interpretability by visualizing activation and attribution
  summarizations,'' \emph{IEEE transactions on visualization and computer
  graphics}, vol.~26, no.~1, pp. 1096--1106, 2019.

\bibitem{lou2012intelligible}
Y.~Lou, R.~Caruana, and J.~Gehrke, ``Intelligible models for classification and
  regression,'' in \emph{Proceedings of the 18th ACM SIGKDD international
  conference on Knowledge discovery and data mining}, 2012, pp. 150--158.

\bibitem{caruana2015intelligible}
R.~Caruana, Y.~Lou, J.~Gehrke, P.~Koch, M.~Sturm, and N.~Elhadad,
  ``Intelligible models for healthcare: Predicting pneumonia risk and hospital
  30-day readmission,'' in \emph{Proceedings of the 21th ACM SIGKDD
  international conference on knowledge discovery and data mining}, 2015, pp.
  1721--1730.

\bibitem{stoica1996synaptic}
A.~Stoica, ``Synaptic and somatic operators for fuzzy neurons: which t-norms to
  choose?'' in \emph{Proceedings of North American Fuzzy Information
  Processing}.\hskip 1em plus 0.5em minus 0.4em\relax IEEE, 1996, pp. 55--58.

\bibitem{jang1993anfis}
J.-S. Jang, ``Anfis: adaptive-network-based fuzzy inference system,''
  \emph{IEEE transactions on systems, man, and cybernetics}, vol.~23, no.~3,
  pp. 665--685, 1993.

\bibitem{cabalar2012some}
A.~F. Cabalar, A.~Cevik, and C.~Gokceoglu, ``Some applications of adaptive
  neuro-fuzzy inference system (anfis) in geotechnical engineering,''
  \emph{Computers and Geotechnics}, vol.~40, pp. 14--33, 2012.

\bibitem{al2016application}
M.~Al-Mahasneh, M.~Aljarrah, T.~Rababah, and M.~Alu’datt, ``Application of
  hybrid neural fuzzy system (anfis) in food processing and technology,''
  \emph{Food engineering reviews}, vol.~8, no.~3, pp. 351--366, 2016.

\bibitem{yao2019using}
H.~Yao, K.~D. Aaronson, L.~Lu, J.~Gryak, K.~Najarian, and J.~R. Golbus, ``Using
  a fuzzy neural network in clinical decision support for patients with
  advanced heart failure,'' in \emph{2019 IEEE International Conference on
  Bioinformatics and Biomedicine (BIBM)}.\hskip 1em plus 0.5em minus
  0.4em\relax IEEE, 2019, pp. 995--999.

\bibitem{sugeno1991successive}
M.~Sugeno and K.~Tanaka, ``Successive identification of a fuzzy model and its
  applications to prediction of a complex system,'' \emph{Fuzzy sets and
  systems}, vol.~42, no.~3, pp. 315--334, 1991.

\bibitem{sun1994rule}
C.-T. Sun, ``Rule-base structure identification in an adaptive-network-based
  fuzzy inference system,'' \emph{IEEE Transactions on Fuzzy Systems}, vol.~2,
  no.~1, pp. 64--73, 1994.

\bibitem{pratama2016incremental}
M.~Pratama, J.~Lu, E.~Lughofer, G.~Zhang, and M.~J. Er, ``An incremental
  learning of concept drifts using evolving type-2 recurrent fuzzy neural
  networks,'' \emph{IEEE Transactions on Fuzzy Systems}, vol.~25, no.~5, pp.
  1175--1192, 2016.

\bibitem{ali2015comparison}
O.~A.~M. Ali, A.~Y. Ali, and B.~S. Sumait, ``Comparison between the effects of
  different types of membership functions on fuzzy logic controller
  performance,'' \emph{International Journal}, vol.~76, pp. 76--83, 2015.

\bibitem{monicka2011performance}
J.~G. Monicka, N.~G. Sekhar, and K.~R. Kumar, ``Performance evaluation of
  membership functions on fuzzy logic controlled ac voltage controller for
  speed control of induction motor drive,'' \emph{International Journal of
  Computer Applications}, vol.~13, no.~5, pp. 8--12, 2011.

\bibitem{sadollah2018fuzzy}
A.~Sadollah, \emph{Fuzzy Logic Based in Optimization Methods and Control
  Systems and Its Applications}.\hskip 1em plus 0.5em minus 0.4em\relax
  BoD--Books on Demand, 2018.

\bibitem{nori2019interpretml}
H.~Nori, S.~Jenkins, P.~Koch, and R.~Caruana, ``Interpretml: A unified
  framework for machine learning interpretability,'' \emph{arXiv preprint
  arXiv:1909.09223}, 2019.

\bibitem{meher2007new}
S.~K. Meher, ``A new fuzzy supervised classification method based on
  aggregation operator,'' in \emph{2007 Third International IEEE Conference on
  Signal-Image Technologies and Internet-Based System}.\hskip 1em plus 0.5em
  minus 0.4em\relax IEEE, 2007, pp. 876--882.

\bibitem{2020SciPy-NMeth}
P.~Virtanen, R.~Gommers, T.~E. Oliphant, M.~Haberland, T.~Reddy, D.~Cournapeau,
  E.~Burovski, P.~Peterson, W.~Weckesser, J.~Bright, S.~J. {van der Walt},
  M.~Brett, J.~Wilson, K.~J. Millman, N.~Mayorov, A.~R.~J. Nelson, E.~Jones,
  R.~Kern, E.~Larson, C.~J. Carey, {\.I}.~Polat, Y.~Feng, E.~W. Moore,
  J.~{VanderPlas}, D.~Laxalde, J.~Perktold, R.~Cimrman, I.~Henriksen, E.~A.
  Quintero, C.~R. Harris, A.~M. Archibald, A.~H. Ribeiro, F.~Pedregosa, P.~{van
  Mulbregt}, and {SciPy 1.0 Contributors}, ``{{SciPy} 1.0: Fundamental
  Algorithms for Scientific Computing in Python},'' \emph{Nature Methods},
  vol.~17, pp. 261--272, 2020.

\bibitem{chang2011libsvm}
C.-C. Chang and C.-J. Lin, ``Libsvm: a library for support vector machines,''
  \emph{ACM transactions on intelligent systems and technology (TIST)}, vol.~2,
  no.~3, pp. 1--27, 2011.

\bibitem{chen2016xgboost}
T.~Chen and C.~Guestrin, ``Xgboost: A scalable tree boosting system,'' in
  \emph{Proceedings of the 22nd acm sigkdd international conference on
  knowledge discovery and data mining}, 2016, pp. 785--794.

\end{thebibliography}

\appendices
\setcounter{figure}{0}    
\setcounter{table}{0}

\section{Implementation of the established ML algorithms}
\label{sec:a1}
Public Python packages were used to build the established ML classifiers with default settings except for specified hyperparameters to be tuned on the validation set.
\begin{enumerate}
  \item Logistic Regression: We used the Logistic Regression Classifier from \textit{sklearn}\cite{2020SciPy-NMeth}.
  \item Na\"ive Bayes: We used Naive Bayes from \textit{sklearn}.
  \item Decision Tree: We used Decision Tree Classifier from \textit{sklearn}. The maximal depth of the tree and the minimum number of samples required to split were tuned.
  \item Random Forest: We used Random Forest Classifier from \textit{sklearn}. The number of trees, the maximal depth of the tree, and the minimum number of samples required to split were tuned.
  \item SVM: We used Support Vector Classifier from \textit{sklearn}, whose implementation is based on \textit{libsvm} \cite{chang2011libsvm}. The regularization parameter, the kernel type (linear function, radial basis function, sigmoid function, or polynomial function), and kernel coefficient were tuned.
  \item XGBoost: We used the tree-based XGBoost Classifier from \textit{xgboost} \cite{chen2016xgboost}. The number of boosting rounds, learning rate, maximal tree depth for base learners were tuned.
  \item EBM: We used the Explainable Boosting Classifier from \textit{interpret}. The learning rate and ways of feature interactions were tuned.
  \item Fuzzy inference classifier: We used Fuzzy Reduction Rule from \textit{fylearn} \href{https://github.com/sorend/fylearn} ({https://github.com/sorend/fylearn}). The classifier used a pi-type membership function and fuzzy mean aggregation \cite{meher2007new}. 
  
\end{enumerate}

\end{document}